\documentclass[runningheads]{llncs}
\usepackage[T1]{fontenc}
\usepackage{graphicx,verbatim}
\usepackage{hyperref}
\usepackage{amsmath,amssymb}
\usepackage{color}

\begin{document}
\title{HD-TTA: Hypothesis-Driven Test-Time Adaptation for Safer Brain Tumor Segmentation}
\titlerunning{HD-TTA for Safer Brain Tumor Segmentation}
%

\author{Kartik Jhawar*, Lipo Wang}  
\authorrunning{Jhawar et al.}
\institute{Institute for Digital Molecular Analytics and Science, Nanyang Technological University, Singapore 636921 \\
School of Electrical and Electronic Engineering, Nanyang Technological University, Singapore 639798\\
    \email{*kartikvi001@e.ntu.edu.sg}}

\maketitle              
\begin{abstract}
Standard Test-Time Adaptation (TTA) methods typically treat inference as a blind optimization task, applying generic objectives to all or filtered test samples. In safety-critical medical segmentation, this lack of selectivity often causes the tumor mask to spill into healthy brain tissue or degrades predictions that were already correct. We propose Hypothesis-Driven TTA, a novel framework that reformulates adaptation as a dynamic decision process. Rather than forcing a single optimization trajectory, our method generates intuitive competing geometric hypotheses: compaction (is the prediction noisy? trim artifacts) versus inflation (is the valid tumor under-segmented? safely inflate to recover). It then employs a representation-guided selector to autonomously identify the safest outcome based on intrinsic texture consistency. Additionally, a pre-screening Gatekeeper prevents negative transfer by skipping adaptation on confident cases. We validate this proof-of-concept on a cross-domain binary brain tumor segmentation task, applying a source model trained on adult BraTS gliomas to unseen pediatric and more challenging meningioma target domains. HD-TTA improves safety-oriented outcomes (Hausdorff Distance (HD95) and Precision) over several state-of-the-art representative baselines in the challenging safety regime, reducing the HD95 by approximately 6.4 mm and improving Precision by over 4\%, while maintaining comparable Dice scores. These results demonstrate that resolving the safety-adaptation trade-off via explicit hypothesis selection is a viable, robust path for safe clinical model deployment. Code will be made publicly available upon acceptance.


\end{abstract}
%
%
%
\section{Introduction}

Deep neural networks have become strong performers for medical image segmentation when the training and test images share similar scanners and acquisition protocols. In brain MRI, robust U-Net style pipelines such as nnU-Net are often a strong starting point \cite{isensee2021nnunet}. However, clinical deployment rarely satisfies the matched-distribution assumption. Differences in scanner vendor, sequence parameters, preprocessing, or patient cohorts can shift image statistics and pathology appearance, leading to missed tumor regions, false positive islands, and boundary leakage. These failures are not only about average overlap, they directly affect whether outputs are reliable for downstream use.

Test-time adaptation (TTA) improves inference robustness without target labels. Approaches range from classic test-time augmentation \cite{shanmugam2021better} to online parameter-update methods using entropy minimization \cite{wang2021tent}, often stabilized by sample filtering \cite{niu2023sar,niu2022eata} or teacher-student frameworks \cite{wang2022ctta}. Recent state-of-the-art methods introduce proxy supervision via self-training \cite{ma2024ist}, feature-statistics alignment \cite{you2025tca}, or segmentation-oriented evaluation objectives \cite{zhou2025tegda}.

Despite success in general vision, many TTA methods remain \emph{blind} to specific medical risks. While some approaches incorporate priors to constrain adaptation \cite{yang2022source,karani2021test}, they often incur high computational overhead or assume fixed strategies that fail on heterogeneous tumors. Recent divide-and-conquer frameworks enable specialized adaptation but rely on interactive user clicks \cite{kim2025dc}, unsuitable for label-free deployment. Crucially, most automated methods still attempt to adapt every / filtered test case(s) using a generic objective. In safety-critical segmentation, this lack of selectivity causes negative transfer: unnecessary updates on easy cases reduce Precision by overfitting noise, while generic objectives under severe shift produce confident but anatomically implausible boundary leakage. Since standard Dice measurement hides these harms, we focus on Hausdorff distance (HD95) to capture boundary risk and Precision for false positive suppression.

\begin{figure}[t]
    \centering
    \includegraphics[width=\textwidth]{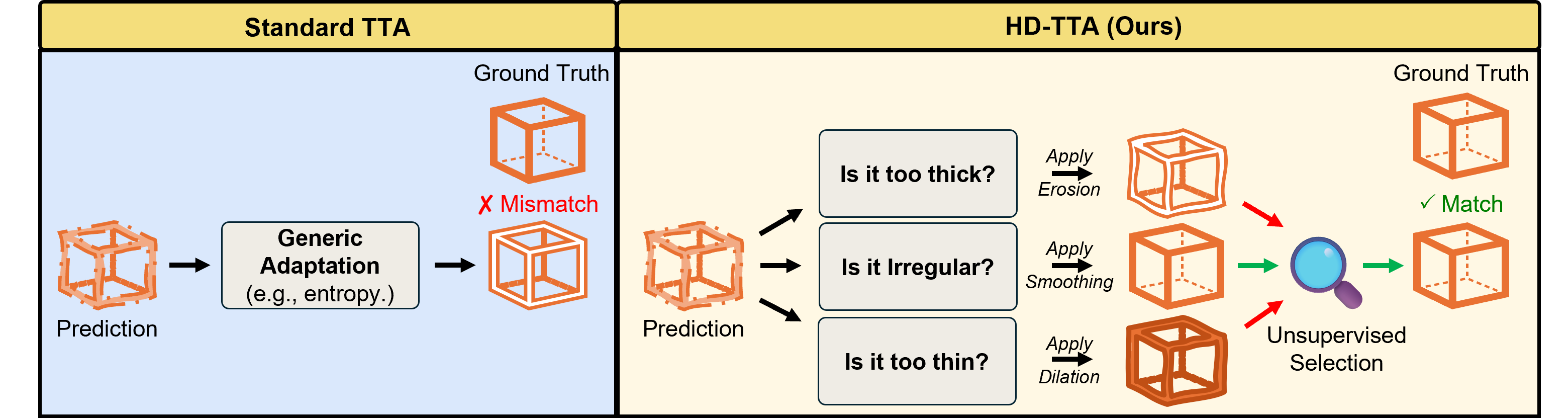}
    \caption{Comparison of Standard TTA versus our proposed HD-TTA framework.}
    \label{fig1:comparison}
\end{figure}
We propose \textbf{Hypothesis-Driven Test-Time Adaptation (HD-TTA)}, a decision-oriented framework that prioritizes \emph{selective} and \emph{safe} refinement rather than unconditional optimization. Given a test volume, HD-TTA produces a baseline prediction, decides whether refinement is warranted, generates a small set of competing geometrically meaningful hypotheses, and selects the safest output using unsupervised plausibility signals (see Fig.~\ref{fig1:comparison}). In short, HD-TTA is the novel unsupervised framework capable of dynamically switching strategies during test time adaptation, preventing the catastrophic failure (negative transfer) that occurs when standard inflation methods are applied to stable cases. 

We validate HD-TTA on the distinct BraTS 2023 Pediatric (PED)~\cite{brats2023PED,brats2023ped_results} and Meningioma (MEN)~\cite{brats2023MEN,brats2023men_results} target-domains using a nnU-Net v2~\cite{isensee2021nnunet} backbone trained on BraTS 2023 GLI (glioma) dataset~\cite{akbari2017segmentation,bakas2017advancing,baid2021rsna,menze2014multimodal,bakas2017segmentationLGG}. Crucially, we apply the unchanged optimization framework with strictly fixed hyperparameters across domains to demonstrate out-of-the-box robustness. Our focus is on demonstrating consistent relative improvements in safety-critical metrics (HD95 and Precision) under distribution shifts, rather than comparison with fully supervised in-domain leaderboard results.

The minimal technical contribution of HD-TTA is hypothesis-conditioned parallel logit refinement combined with unsupervised hypothesis selection under frozen backbone parameters. The specific proxy losses used for compactness and diffusion are instantiations of this framework and can be replaced. The Gatekeeper and selector are modular components. While this work focuses on binary tumor segmentation, the hypothesis-driven refinement framework is general and can be extended to multi-class or other structured prediction tasks.

\section{Method}
\label{sec:method}

\begin{figure}[t]
    \centering
    \includegraphics[width=\textwidth]{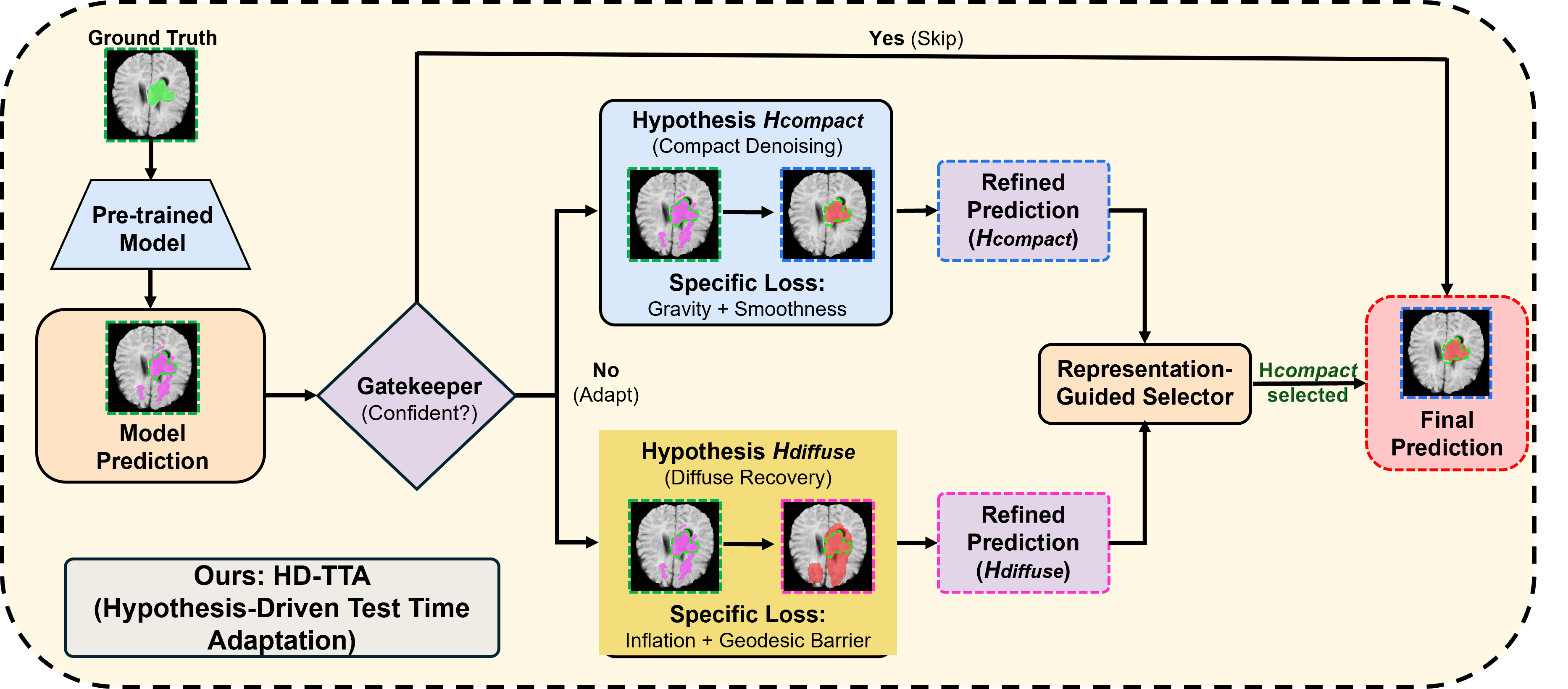}
    \caption{Overview of the proposed HD-TTA framework. The inference pipeline consists of a selective Gatekeeper, parallel hypothesis generation (e.g., $H_{compact}$ winning to trim spurious false positives, versus $H_{diffuse}$), and an unsupervised representation-guided selector.}
    \label{fig2:pipeline}
\end{figure}

As illustrated in Fig.~\ref{fig2:pipeline}, HD-TTA operates as an orthogonal refinement layer on top of a pre-trained (on BraTS-GLI) segmentation network nnU-Net v2 (backbone). We consider binary brain tumor segmentation from a 3D multi-modal MRI volume. Given a target-domain test case $x$, its initial backbone prediction $P_0$, and its corresponding initial logits $z_0$. During adaptation, optimization is performed directly on the actively updating test-sample logits $z$ while keeping all backbone parameters completely frozen; the current probabilities $P = \text{sigmoid}(z)$ are computed at each step solely for loss evaluation and final mask generation. The framework executes three sequential stages: (1) a selective Gatekeeper to prevent unnecessary updates, (2) the parallel generation of competing geometric Hypotheses, and (3) a Representation-Guided Selection step to identify the safest outcome. 

\subsection{Selective Adaptation via the Gatekeeper}
To mitigate negative transfer on already accurate predictions, a lightweight Gatekeeper assesses the stability of $P_0$. It flags a sample for hypothesis-conditioned refinement if its volume is critically small ($<300$ voxels, risking for tumor collapse) or if its uncertainty ratio (proportion of the predicted tumor volume where $0.3 < P_0 < 0.7$) exceeds 5\%. Confident samples skip adaptation.


\subsection{Hypothesis-Conditioned Refinement}
Two intuitive competing hypotheses for the proof-of-concept binary brain tumor segmentation task are as follows:

\textbf{Hypothesis 1: Compact Denoising ($H_{\text{compact}}$).}
This hypothesis addresses over-segmentation and spurious noise islands. A loss function that enforces smoothness and spatial compactness:
\begin{equation}
    \mathcal{L}_{compact} = \lambda_{ent} \mathcal{H}(P) + \underbrace{\lambda_{TV} ||\nabla P||_1 + \lambda_{grav} \mathcal{V}(P)}_{\textbf{Specific to } H_{compact}} + \lambda_{anc} ||z - z_0||^2
\end{equation}
Here, $\lambda$ represents the weighting coefficients,  $\mathcal{H}(P)$ is the entropy, and $||\nabla P||_1$ is standard Total Variation (TV) to smooth jagged borders \cite{rudin1992tv}. The crucial term is the gravity loss $\mathcal{V}(P)$, which minimizes the spatial variance of the foreground coordinates, effectively pulling outlier pixels toward the tumor centroid. An anchor term penalizes deviation from the baseline logits $P_0$.

\textbf{Hypothesis 2: Diffuse Recovery ($H_{\text{diffuse}}$).}
This hypothesis addresses under-segmentation by encouraging controlled growth. To prevent uncontrolled leakage into healthy tissue, we replace standard TV with a geodesic barrier \cite{caselles1997geodesic}:
\begin{equation}
    \mathcal{L}_{diffuse} = \lambda_{ent} \mathcal{H}(P) + \underbrace{\lambda_{geo} \sum (g \cdot |\nabla P|) + \lambda_{inf} (-\mu_P)}_{\textbf{Specific to } H_{diffuse}} + \lambda_{anc} ||z - z_0||^2
\end{equation}
The inflation term ($-\mu_P$) drives expansion, constrained by a geodesic barrier $g$ derived from image gradients. Expansion halts at anatomical edges where $g \approx 0$, preventing leakage into the skull or ventricles. 
\subsection{Representation-Guided Selection}
To autonomously select the safest hypothesis without ground truth, we utilize an intrinsic texture consistency signal. This is a proof-of-concept selector; however, more test-sample-only perturbation stability-based or morphology-based selectors can be swapped in without changing the overall HD-TTA framework. 
Since blind expansion carries a high risk of boundary leakage (increasing HD95), we treat $H_{\text{compact}}$ as the safe default and require $H_{\text{diffuse}}$ to pass a strict texture validation check to be selected. We define the high-confidence "tumor core" ($P_0 > 0.8$) and the expansion candidate region $\Delta = H_{diffuse} \setminus \text{Core}$(i.e., the pixels added by the inflation hypothesis). We evaluate the consistency of this candidate region against the core:
\begin{equation}
    S_{\text{rep}} = \exp \left( -0.5 \cdot \left( \frac{|\mu_{\Delta} - \mu_{\text{core}}|}{\gamma (\sigma_{\text{core}} + \epsilon)} \right)^2 \right)
\end{equation}
where $\mu$ and $\sigma$ denote the mean and standard deviation of intensity, and $\gamma = 1.5$ serves as a tolerance scaling factor. The model selects $H_{\text{diffuse}}$ only if the newly recruited pixels share the intrinsic intensity signature of the core ($S_{\text{rep}} > 0.95$); otherwise, it reverts to the safe default, $H_{\text{compact}}$. 
\footnote{All hyperparameters, including loss coefficients in Section 2.2 ($H_{compact}$: $\lambda_{ent}=10$, $\lambda_{TV}=0.5$, $\lambda_{grav}=50$, $\lambda_{anc}=50$; $H_{diffuse}$: $\lambda_{ent}=2$, $\lambda_{geo}=50$, $\lambda_{inf}=5$, $\lambda_{anc}=0.1$), Gatekeeper thresholds in Section 2.1, and the Selection thresholds in Section 2.3, were established during initial method development to ensure stable optimization and define safe operational boundaries. Crucially, they were kept strictly fixed for all experiments without any target-domain tuning.}



\section{Experiment and Results}

\begin{table}[t]
\caption{Comparison of HD-TTA against state-of-the-art TTA methods on BraTS-PED and BraTS-MEN target sets. Values are Mean $\pm$ Std. Best values are in \textbf{bold}. $^{\#}$ denotes statistically significant improvement ($p<0.05$) over the best existing method (TCA) via a one-sided paired Wilcoxon signed-rank test with Holm–Bonferroni correction. In our nnU-Net v2 configuration, TEGDA produced unchanged output masks relative to Classic TTA; we therefore report identical metrics and discuss this observation in the text.}
\label{tab:main_results}
\centering
\resizebox{\textwidth}{!}{%
\begin{tabular}{|l|ccc|ccc|}
\hline
 & \multicolumn{3}{c|}{\textbf{BraTS-PED}} & \multicolumn{3}{c|}{\textbf{BraTS-MEN}} \\
\textbf{Method} & \textbf{Dice (\%)} $\uparrow$ & \textbf{HD95 (mm)} $\downarrow$ & \textbf{Prec (\%)} $\uparrow$ & \textbf{Dice (\%)} $\uparrow$ & \textbf{HD95 (mm)} $\downarrow$ & \textbf{Prec (\%)} $\uparrow$ \\
\hline
\multicolumn{7}{|c|}{\textit{No TTA (anchor)}} \\
\hline
nnU-Net v2(2021)\cite{isensee2021nnunet} & 86.47 $\pm$ 11.81 & 8.20 $\pm$ 12.33 & 82.90 $\pm$ 15.58 & 12.81 $\pm$ 22.15 & 73.50 $\pm$ 31.47 & 13.77 $\pm$ 24.95 \\
\hline
\multicolumn{7}{|c|}{\textit{Classic aggregation}} \\
\hline
Classic TTA (2021)\cite{shanmugam2021better} & 86.95 $\pm$ 11.56 & 6.59 $\pm$ 9.75 & 83.44 $\pm$ 15.44 & 13.20 $\pm$ 22.96 & 72.19 $\pm$ 32.42 & 14.63 $\pm$ 26.41 \\
\hline
\multicolumn{7}{|c|}{Entropy-based} \\
\hline
SAR (2023)\cite{niu2023sar} & 85.89 $\pm$ 12.35 & 7.58 $\pm$ 10.35 & 81.12 $\pm$ 16.26 & 12.98 $\pm$ 22.43 & 73.44 $\pm$ 31.77 & 13.41 $\pm$ 24.38 \\
\hline
\multicolumn{7}{|c|}{\textit{Learning-based}} \\
\hline
IST (2024)\cite{ma2024ist} & 86.19 $\pm$ 15.20 & 7.19 $\pm$ 12.88 & 85.33 $\pm$ 15.00 & 2.51 $\pm$ 10.90 & 96.00 $\pm$ 16.15 & 3.01 $\pm$ 13.68 \\
\hline
\multicolumn{7}{|c|}{\textit{Feature alignment}} \\
\hline
TCA (2025)\cite{you2025tca} & \textbf{87.43 $\pm$ 11.13} & 6.39 $\pm$ 9.58 & 84.62 $\pm$ 14.97 & \textbf{13.26 $\pm$ 23.16} & 70.96 $\pm$ 32.48 & 15.36 $\pm$ 27.66 \\
\hline
\multicolumn{7}{|c|}{\textit{Recent adaptive TTA}} \\
\hline
TEGDA (2025)\cite{zhou2025tegda} & 86.95 $\pm$ 11.62 & 6.59 $\pm$ 9.80 & 83.44 $\pm$ 15.52 & 13.20 $\pm$ 23.04 & 72.19 $\pm$ 32.53 & 14.63 $\pm$ 26.50 \\
\hline
\multicolumn{7}{|c|}{\textit{Ours}} \\
\hline
\textbf{HD-TTA} & 87.26 $\pm$ 11.33 & \textbf{5.35 $\pm$ 6.91} & \textbf{86.32 $\pm$ 13.34}$^{\#}$ & 10.57 $\pm$ 19.47 & \textbf{64.55 $\pm$ 30.51}$^{\#}$ & \textbf{19.64 $\pm$ 35.22}$^{\#}$ \\
\hline
\end{tabular}%
}
\end{table}
\textbf{Implementation Details.} We use the BraTS 2023 GLI (official training split, 1251 cases) as the source domain to train a strong base segmenter nnU-Net v2 backbone for 250 epochs with Dice loss and a deterministic polynomial learning-rate decay schedule. For proof-of-concept of the HD-TTA principle, the binary tumor segmentation task, we merge the standard multi-class labels into a single binary whole tumor mask for both training and evaluation. For target domains, we use unchanged optimization procedures on the public 99-case BraTS 2023 PED set and a fixed 144-case BraTS 2023 MEN, randomly sampled from the training split. All methods adapt only predicted logits, keeping backbone weights frozen. Classical TTA \cite{shanmugam2021better}, representing standard test-time augmentation is implemented here natively via nnU-Net's mirroring, which is uniformly active to the initial baseline predictions of all other methods to ensure a fair comparison. For HD-TTA, each borderline case is optimized using two hypotheses-based loss functions in parallel using the Adam optimizer (lr=0.1, 1000 steps). We adapted baselines to use instance-based architectures to match nnU-Net's Instance Normalization. All experiments ran on an NVIDIA RTX A5000 GPU (CUDA 12.2) without target labels. On borderline cases, HD-TTA performs 1000 steps of logit optimization (Adam optimizer, lr=0.1) per hypothesis. On our hardware this corresponds to an average additional latency of 21.91 seconds (BraTS-PED) and 21.89 seconds (BraTS-MEN) per flagged case. The Gatekeeper flags approximately 23.6\% and 99.3\% of cases, resulting in an average overhead of 5.17 seconds and 21.74 seconds per case, respectively. Evaluation metrics include volume-level Dice similarity, the 95th percentile Hausdorff distance (HD95), and precision. 

\textbf{Comparison with several SOTA TTA methods.}
We compare HD-TTA against representative state-of-the-art methods spanning major families of test-time adaptation: (i) \textbf{Classical TTA}, which averages predictions across augmented views; (ii) \textbf{SAR} \cite{niu2023sar}, an entropy-minimization method that filters high-gradient samples; (iii) \textbf{IST} \cite{ma2024ist}, a recent self-training approach; (iv) \textbf{TCA} \cite{you2025tca}, which aligns feature statistics; and (v) \textbf{TEGDA} \cite{zhou2025tegda}, a recent test-time evaluation-guided adaptive method. These methods cover the common design space of TTA, ranging from purely averaging predictions to online adaptation using proxy supervision. 

\textbf{Quantitative Analysis: Safety and Stability.}
Table~\ref{tab:main_results} summarizes the quantitative results across both the BraTS-PED and BraTS-MEN regimes. Across both, HD-TTA consistently improved the safety-oriented objectives. On the stable BraTS-PED dataset, standard nnU-Net already performs strongly (Dice $86.47\%$). Here, aggressive adaptation is often detrimental. Blind optimization methods like SAR and IST fail to improve performance on borderline test cases, likely due to optimization instability on binary tasks where background dominance biases entropy minimization. IST can be vulnerable when pseudo-supervision is unreliable, which can amplify errors under severe shift \cite{ma2024ist}. While the strongest competitor, TCA improved Dice (best on both datasets) by aligning feature statistics, its global alignment objective did not consistently minimize worst-case boundary risk (HD95 $6.39$ mm). In contrast, HD-TTA achieves the lowest boundary error (HD95 $5.35$ mm, a 16.3\% improvement over TCA) and high Precision ($86.32\%$, $+1.7\%$ over TCA), while maintaining comparable Dice. This also confirms that our Gatekeeper successfully prevents negative transfer on easy cases, a capability missing in "always-adapt" baselines. While high standard deviations are observed across all methods-reflecting the intrinsic heterogeneity of the target cohorts as also seen in recent benchmarks like TEGDA \cite{zhou2025tegda}, the primary strength of HD-TTA lies in its consistent directional improvement. Unlike baselines that trade off metrics, HD-TTA consistently moves both safety metrics in the intended direction (lowering HD95 and raising Precision) across diverse datasets.

The generalizability of HD-TTA is pronounced when applied without modifications on the challenging BraTS-MEN dataset, where domain gap is significantly larger with BraTS-GLI due to different tumor location, growth pattern, and contrast characteristics. Thus, Dice scores are low across methods in this cross-domain setting due to severe distribution shift and frequent small or fragmented tumor regions, making false-positive suppression and boundary robustness (Precision and HD95) more informative for safety-oriented evaluation. Most baselines struggle to recover the missed tumor, yielding high HD95 scores ($>70$ mm). HD-TTA significantly improves over the strongest competitor (TCA) in safety metrics, reducing HD95 by approximately $6.4$ mm ($70.96 \to 64.55$ mm) and improving Precision by over $4\%$ ($15.36 \to 19.64\%$). Across both datasets, TEGDA \cite{zhou2025tegda} results are effectively identical to classical TTA: although the adaptation step updates parameters (logit/optimization activity observed), the nnU-Net v2 backbone predictor outputs remain unchanged, indicating the update does not reach the inference graph used to generate the saved predictions. Although the Dice scores remain low across all methods due to the extreme difficulty of this target-domain, HD-TTA's superior Precision and boundary adherence indicate that it effectively suppresses false positives while recovering valid tumor regions, aligning with the safety-critical goals of clinical deployment.

\begin{figure}[t]
    \centering
    \includegraphics[width=\textwidth]{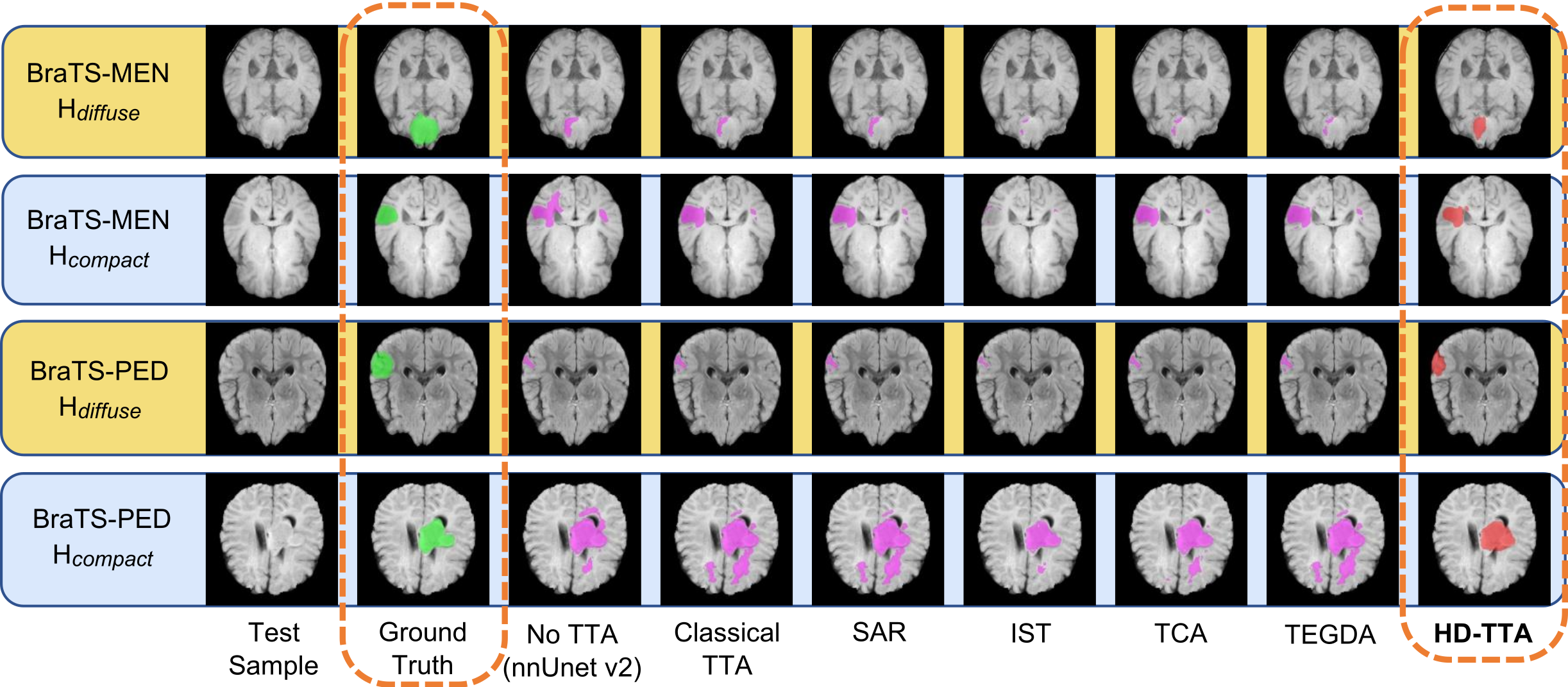}
    \caption{Qualitative comparison on representative BraTS-MEN (Rows 1-2) and BraTS-PED cases (Rows 3-4).}
    \label{fig3:comparison}
\end{figure}

\textbf{Qualitative Mechanisms and Visual Analysis.}
Fig.~\ref{fig3:comparison} provides qualitative evidence for why the safety gains occur. \textbf{\textit{Adaptation on representative BraTS-MEN cases (Fig.~\ref{fig3:comparison}, Rows 1-2):}} The backbone nnU-Net v2 prediction (marked as No TTA) misses significant portions of the tumor (in magenta color). Entropy-based methods (SAR, TCA) fail to recover this missed tissue because they lack a mechanism to initiate growth in low-confidence regions. In Row 1, HD-TTA successfully selects $H_{\text{diffuse}}$, leveraging the inflation objective to relatively recover the missed tumor volume (red). Conversely, in Row 2, HD-TTA correctly selects $H_{\text{compact}}$ to denoise the primary nnU-Net prediction and cleanly eliminate the spurious false-positive island; notably, even the strongest baseline, TCA, fails to fully suppress this artifact. \textbf{\textit{Adaptation on representative BraTS-PED cases (Fig.~\ref{fig3:comparison}, Rows 3-4):}} In Row 3, the backbone nnU-Net v2 under-segments the tumor, prompting HD-TTA to correctly select $H_{\text{diffuse}}$ to recover the missing boundary, when other baselines fail to recover to a similar extent. Conversely, in Row 4, the backbone nnU-Net v2 produces a severe over-segmentation with a prominent spurious noise island. Blind adaptation methods either preserve or exacerbate these high-confidence artifacts (as seen with Classical TTA). HD-TTA's Gatekeeper flags the instability and selects $H_{\text{compact}}$, utilizing gravity and smoothness constraints to cleanly eliminate the false positives and restore a plausible tumor shape. Notably, while IST attempts to suppress this artifact, it fails to remove it entirely.

Overall, the quantitative and qualitative evidence consistently supports the central claim of HD-TTA: rather than blindly optimizing every test case under a single objective, flagging borderline cases and selectively choosing between competing geometric hypotheses leads to safer boundary behavior. While Dice remains comparable to strong baselines, HD-TTA consistently improves boundary reliability (HD95) and false-positive control (Precision), which are critical in safety-sensitive medical segmentation.

\textbf{Ablation Study.} Table~\ref{tab:ablation} details the impact of each component on the BraTS-PED dataset.
First, regarding hypothesis selection, we observe that forcing the inflation strategy (only $H_{\text{diffuse}}$) is dangerous on stable cases, causing significant boundary leakage (HD95 spikes to $9.31$ mm). In contrast, Full HD-TTA successfully identifies this risk and converges to the performance of the safe $H_{\text{compact}}$ strategy (HD95 $5.35$ mm), proving that the selector correctly rejects unsafe updates.
Second, the gatekeeper is essential for preventing negative transfer. Blindly adapting every case (w/o Gatekeeper) forces $H_{compact}$ onto already-accurate predictions. This aggressive boundary erosion artificially inflates Precision by eliminating false positives, but simultaneously degrades overall overlap (Dice 85.38\%) and boundary accuracy (HD95 6.19 mm) by deleting valid tumor tissue. This trade-off is specific to the high-stability BraTS-PED baseline and does not diminish the necessity of HD-TTA. Also, as indicated, the Gatekeeper is critical to prevent unnecessary adaptation on confident cases, while hypothesis selection primarily acts as a safeguard against unstable inflation. On relatively stable domains such as BraTS-PED, safety improvements arise mainly from selective suppression rather than frequent inflation. Finally, removing the edge map constraint worsens boundary adherence (HD95 increases to $5.44$ mm), confirming its role as a necessary physical guardrail against leakage.

\begin{table}[t]
\caption{Ablation study on the BraTS-PED dataset. Values are Mean $\pm$ Std. Best values are in \textbf{bold}.}
\label{tab:ablation}
\centering
\begin{tabular}{|l|ccc|}
\hline
\textbf{Method} & \textbf{Dice (\%)} $\uparrow$ & \textbf{HD95 (mm)} $\downarrow$ & \textbf{Prec (\%)} $\uparrow$ \\
\hline
w/o Gatekeeper & 85.38 $\pm$ 10.91 & 6.19 $\pm$ 6.87 & \textbf{90.19 $\pm$ 14.15} \\
w/o Edge Map & 87.22 $\pm$ 11.43 & 5.44 $\pm$ 7.08 & 86.70 $\pm$ 12.86 \\
\hline
only $H_{\text{compact}}$ & 87.26 $\pm$ 11.33 & 5.35 $\pm$ 6.97 & 86.32 $\pm$ 13.34 \\
only $H_{\text{diffuse}}$ & 82.18 $\pm$ 18.17 & 9.31 $\pm$ 13.21 & 76.83 $\pm$ 23.20 \\
\hline
\textbf{Full HD-TTA} & \textbf{87.26 $\pm$ 11.38} & \textbf{5.35 $\pm$ 6.95} & 86.32 $\pm$ 13.40 \\
\hline
\end{tabular}
\end{table}

\section{Conclusion}
We presented HD-TTA, a decision-oriented inference framework that replaces blind per-case optimization with selective refinement and hypothesis-conditioned updates. In short, HD-TTA operates on a clear directive: flag borderline cases, hypothesize tailored strategies, and autonomously select the "safest" correction. Using a strong nnU-Net v2 backbone trained on BraTS 2023 GLI, we evaluated HD-TTA without any target-domain tuning on BraTS-PED and more challenging BraTS-MEN. We observed consistent improvements in safety-oriented metrics, achieving lower boundary error (HD95) and higher Precision relative to representative TTA baselines, while maintaining comparable Dice. These results along with the qualitative analysis support the core principle that explicitly structured, competing refinement hypotheses can better control failure modes such as under-segmentation, boundary leakage and spurious islands under domain shift.

\textbf{Acknowledgments.} 
This research is supported by the Ministry of Education, Singapore, under its Research Centre of Excellence award to the Institute for Digital Molecular Analytics \& Science, NTU (IDMxS, grant: EDUNC-33-18-279-V12)

\textbf{Disclosure of Interests.}
The authors declare no conflicts of interest.
%
%
%
\bibliographystyle{splncs04}
\bibliography{refs}




\end{document}